\documentclass[runningheads]{llncs}
\usepackage{hyperref}
\usepackage{graphicx}
\usepackage{booktabs}
\usepackage{tikz}
\usepackage{multirow}
\usetikzlibrary{calc}
\newcommand*\circled[1]{\tikz[baseline=(char.base)]{
            \node[shape=circle,draw,minimum size=10pt,inner sep=0pt] (char) {\scriptsize{\textbf{#1}}};}}

\begin{document}
\title{Machine Learning for Fraud Detection in E-Commerce: A Research Agenda}
%
%\titlerunning{Abbreviated paper title}
% If the paper title is too long for the running head, you can set
% an abbreviated paper title here
%

\author{Niek Tax \and
Kees Jan de Vries \and
Mathijs de Jong \and
Nikoleta Dosoula \and
Bram van den Akker \and Jon Smith \and Olivier Thuong \and
Lucas Bernardi}
\authorrunning{N. Tax et al.}
% First names are abbreviated in the running head.
% If there are more than two authors, 'et al.' is used.
%
\institute{Booking.com, Amsterdam, The Netherlands
\email{\{niek.tax,kees.devries,mathijs.dejong,nikoleta.dosoula,\\bram.vandenakker,jon.smith,olivier.thuong,lucas.bernardi\}@booking.com}}
\maketitle              % typeset the header of the contribution
\begin{abstract}
Fraud \emph{detection} and \emph{prevention} play an important part in ensuring the sustained operation of any e-commerce business. \emph{Machine learning} (ML) often plays an important role in these anti-fraud operations, but the organizational context in which these ML models operate cannot be ignored. In this paper, we take an organization-centric view on the topic of fraud detection by formulating an \emph{operational model} of the anti-fraud departments in e-commerce organizations. We derive 6 research topics and 12 practical challenges for fraud detection from this operational model. We summarize the state of the literature for each research topic, discuss potential solutions to the practical challenges, and identify 22 open research challenges.

\end{abstract}
\section{Introduction}
E-commerce is an important and rapidly growing sector that has tripled its share of the world GDP from 0.5\% to $>$1.5\% in the past decade~\cite{Moagar2017}. This surge in economic importance is accompanied by a rapid increase in the total cost of global cybercrime, which increased from \$445 billion in 2014 to $>$\$600 billion in 2017~\cite{McAfee2018}. Fraud and cybercrime in the e-commerce domain spans a variety of fraud types, such as \emph{fake accounts}~\cite{Cavico2017}, \emph{payment fraud}, \emph{account takeovers}~\cite{Kawase2019}, and \emph{fake reviews}.

\emph{Machine learning} (ML) plays an important role in the detection, prevention, and mitigation of fraud in e-commerce organizations. Publicly-known examples include Microsoft~\cite{Nanduri2020}, LinkedIn~\cite{Wang2020}, and eBay~\cite{Porwal2019}. In practice, fraud detection ML models in e-commerce organizations do not operate in isolation, but they are embedded in a larger \emph{anti-fraud department} that also employs \emph{fraud analysts} or \emph{fraud investigators} who perform case investigations and proactively search for fraud trends. This requires fraud detection models to be embedded in the way of working and daily operations of an anti-fraud department.

While the existing literature on fraud detection is extensive, to the best of our knowledge there is currently no work that provides an explicit formulation of the daily operations of anti-fraud departments. This creates a gap between academic work on fraud detection and practical applications of fraud detection in industry. Furthermore, this makes it more difficult to assess whether novel fraud detection methods fit into the practical way-of-working in fraud departments, or whether they address practically relevant challenges.

In this paper, we describe the operational model of an anti-fraud department. We use this operational model to derive a set of practically relevant research topics for fraud detection. For each research topic, we summarize the state of the literature and put forth a set of \emph{open research challenges} that are formulated from a practical angle. The main aim of this paper is to put forward a research agenda of open challenges in fraud detection.

This paper is structured as follows. In~\autoref{sec:operational_model} we introduce and discuss the operational model of fraud detection from an organizational point of view, discuss the role that machine learning plays in it, and derive research topics from it. In~\autoref{sec:presentation} to~\autoref{sec:model_training}, we zoom in on each of the individual research topics that we introduce in~\autoref{sec:operational_model}. In each of those sections, we discuss one research topic, list \emph{practical considerations from industry experience}, \emph{summarize the current state of the literature}, and \emph{formulate open research challenges}. We conclude this paper in~\autoref{sec:conclusion}. 

\section{An Operational Model of an Anti-Fraud Department}
\label{sec:operational_model}
\begin{figure}[t]
    \centering
    \includegraphics[width=\textwidth]{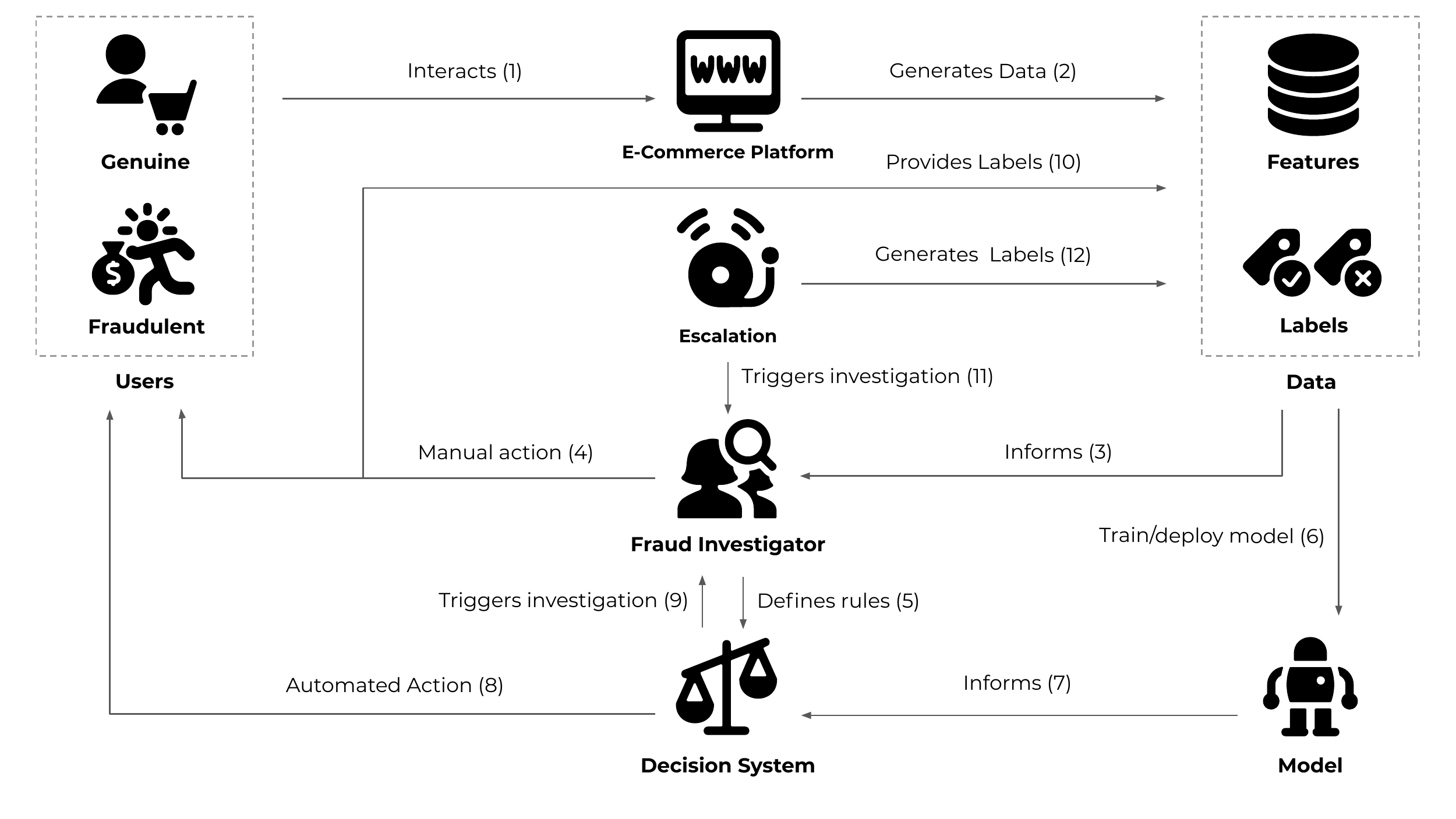}
    \vspace{-0.7cm} 
    \caption{A model of the daily operations of an anti-fraud department in an e-commerce organization.}
    \label{fig:operational_model}
\end{figure}

In this section, we introduce our operational model~(\autoref{fig:operational_model}) of the daily operations of anti-fraud departments in e-commerce organizations. We highlight the role of machine learning in the daily operations of anti-fraud departments and derive research topics and practical challenges.

\subsection{Entities \& Relations in the Operational Model}

\begin{description}
    \item[E-Commerce Platform]{Online service where \emph{users} can buy
    and sell products (e.g., Amazon, Booking.com, or Zalando).}
    \item[Users]{
    \emph{Genuine users} perform legitimate transactions (e.g., purchases or sales) on the \emph{e-commerce platform}.
    \emph{Fraudulent users} are wrongful or criminal actors who intend to achieve financial or personal gains through fraudulent activity on the \emph{e-commerce platform}. Examples of such fraudulent activity: purchase attempts with stolen credit cards, abuse of marketing initiatives (e.g., incentive programs), registering fake accounts (e.g. merchant accounts or user accounts), phishing, or other attempts at account take-overs.
    Users interact with the \emph{e-commerce platform} \circled{1}, which in turn, generates \emph{data} \circled{2}.
    }
    \item[Data]{Is generated by the \emph{e-commerce platform} as a result of user interactions. From the ML viewpoint, data can be transformed into \emph{features} and \emph{labels}. Features represent relevant behavior (e.g., browsing, purchasing, messaging, or managing accounts), or business entities (e.g., purchases, products, or users) of fraudulent and legitimate users. 
    Labels indicate whether or not behavior or an entity is fraudulent.
    Labels often result from investigation results of \emph{fraud investigators} \circled{10}.
    Sometimes, labels arrive through \emph{external escalations} \circled{12}, e.g., through notifications of fraud from credit card issuers.}
    \item[Fraud Investigator]{
    Professionals who investigate suspected fraud cases, using the \emph{data} \circled{3}. These suspected fraud cases might originate from internal escalations \circled{11} (e.g., complaints through customer service), or the \emph{decision  system}  triggered an investigation \circled{9}.
    For fraud that they find they take \emph{remediating actions} \circled{4} (e.g., canceling orders or blocking users) and/or 
    \emph{preventative steps} by defining \emph{rules} \circled{5} that are aimed to identify similar fraud in the future, which are used in the \emph{decision system}.
    }
    % \item[ML Practitioner]{The collective of ML/Data Scientists, and ML/Software Engineers who are responsible for developing Fraud Detection models, as well as their deployment.}
    \item[Decision System]{
    A system that can take concrete actions for instances.
    \emph{Instances} arise from specific \emph{user requests}, e.g., the purchase of a product, or registration of an account.
    Instances require a decision on what action to take, e.g., \emph{no intervention}, \emph{to request additional verification}, or to fully \emph{block} the user's request. The decision system can take automatic action \circled{8}, or trigger an investigation by a \emph{fraud investigator} \circled{9} (who could then take manual action through \circled{4}).
    Actions can either be \emph{synchronous} (i.e., blocking the user request) or \emph{asynchronous} (i.e., without blocking the user request).
    The decision system decides on its actions by combining \emph{ML models} \circled{7} and \emph{rules} \circled{5}.
    In addition, some use cases require \emph{exploration}, e.g., by occasionally triggering investigations on instances where there is high uncertainty on whether they are fraudulent. 
    }
    \item[Model]{A machine learning model that aims to distinguish between fraudulent and genuine users. This model is trained \circled{6} on the \emph{data}.
    }
\end{description}

\subsection{Research Topics}
% We discuss the following research topics in the remainder of this paper, and highlight how they relate to~\autoref{fig:operational_model}.
We now discuss the \emph{research topics} that arise from~\autoref{fig:operational_model}. These \emph{research topics} form the basis of the remainder of this paper, where we dedicate one section per research topic, list their concrete \emph{practical challenges}, provide a summary of existing \emph{solution areas in the literature}, and identify \emph{open research challenges}.

\autoref{tab:research_topics} summarizes all research topics, their connection to~\autoref{fig:operational_model}, their practical challenges, and the solution areas in the literature. These connections are either a \emph{set of edges}, or a \emph{path of edges} in~\autoref{fig:operational_model}. In the latter case, $\Rightarrow$(\circled{X}, \circled{Y}) denotes a path consisting of edges $X$ and $Y$. Below we introduce the research topics and highlight the practical challenges in \textbf{bold}.

\begin{table}[t]
    \centering
    \resizebox{\textwidth}{!}{
    \begin{tabular}{|l|l|l|l|}
        \toprule
        \textbf{Research topic} & \textbf{Connection to~\autoref{fig:operational_model}} & \textbf{Practical challenges}&\textbf{Solutions areas}\\
        
        \midrule
        Investigation support & \circled{3}, \circled{9}, \circled{11} & Limited capacity & Explainable AI\\ 
        && Outcome accuracy & Multiple instance learning\\ &&Grouped investigations & Network learning\\
        \midrule
        Decision-making & \circled{5}, \circled{7}, \circled{8}, \circled{9} & Risk management  & Probability calibration\\
        &&Consequential actions& Cost-sensitive learning\\ 
        &&Combining rules \& ML& AI fairness\\
        &&& Rule-based systems\\ 
        && & Uncertainty quantification\\
        \midrule
        Labels & \circled{8}, \circled{10}, \circled{12} & Selection bias & Learning under selection bias \\
        & & & Multi-armed bandits\\
        \midrule
        Concept drift & $\Rightarrow$(\circled{1}, \circled{2}, \circled{6}, \circled{7}, \circled{8}) & Adversarial drift & Concept drift adaptation\\
        &$\Rightarrow$(\circled{1}, \circled{2}, \circled{6}, \circled{7}, \circled{9}, \circled{4})& Natural drift & Adversarial robustness\\
        && Upstream models & Anomaly detection\\
        % && Delayed labels & \\
        %&&  & Delayed labels\\
        \midrule
        ML model--investigator & $\Rightarrow$(\circled{6}, \circled{7}, \circled{9}, \circled{10}) & Explore/exploit trade-off & Active learning\\
        interaction& & & Guided learning\\
        & & & Weak supervision\\
        %& & & Human-in-the-loop ML\\
        \midrule
        Model & \circled{6} & Model reliability & Model verification\\
        %&& & Feature engineering\\ 
        && & Automated data validation\\
        && & Deployment best practices\\
        \bottomrule
    \end{tabular}
    }
    \caption{An overview of the \emph{research topics} that we derived from~\autoref{fig:operational_model}, the \emph{practical challenges}, and \emph{solution areas} in the literature that relate to them.}
    \label{tab:research_topics}
    \vspace{-0.2cm}
\end{table}

\begin{description}
    \item[Investigation support]{
    Fraud investigations (triggered by \circled{9} or \circled{11}) are performed based on evidence from the data \circled{3}. \emph{Fraud investigators} have \textbf{limited time capacity}, and to avoid alarm fatigue, \circled{9} and \circled{11} must yield a high precision. Furthermore, investigators must be enabled and supported to reach decisions \emph{efficiently} and \textbf{accurately}.
    There are several opportunities for machine learning to play a role in supporting these investigations and in the evidence-gathering process that they entail. For example, some fraud cases are highly similar because they are part of the same attack (e.g., they might be performed from the same IP address). Ideally, these are \textbf{grouped into a single investigation}, to minimize the number of investigations, and to provide context to the \emph{fraud investigator} during the investigation.
    We summarize and discuss this research topic in~\autoref{sec:presentation}.}
    \item[Decision-making]{Relations \circled{5}, \circled{7}, \circled{8}, and \circled{9} show the role of the \emph{decision system}, which is tasked to decide \emph{which} instances to take action on by \textbf{combining the output of the ML model and rules} that were created by the \emph{fraud investigators}. The \emph{decision system} is also tasked to decide \emph{how} to take action: either \emph{automatically and immediately} or by sending the case for further review to \emph{fraud investigators}. These decisions should be made with the aim to \textbf{manage risk}, i.e., possible risks of negatively impacting genuine users should be traded-off with the possible risk of failing to block fraud.
    The preventative or remediating actions (e.g., disabling accounts, or stopping purchases) have \textbf{great consequences} to the user by design. Therefore, it is essential to limit false positives and to take fairness into account. 
    In~\autoref{sec:decision_making}, we discuss how the research areas of \emph{cost-sensitive learning}, \emph{AI fairness}, and \emph{uncertainty quantification} offer partial solutions to these challenges and formulate remaining open challenges.}
    \item[Labels]{The two sources of labels are \emph{fraud investigators} \circled{10}, and automatic escalations \circled{12}. These mechanisms introduce \textbf{selection bias} through delay or incompleteness of labels. In addition, automated actions \circled{8} that block suspected fraud mask labels from automatic escalations. We discuss in~\autoref{sec:feedback} how \emph{learning under selection bias} and  \emph{multi-armed bandits} offer solutions and we formulate open challenges.}
    \item[Concept drift]{The cycles $\Rightarrow$(\circled{1}, \circled{2}, \circled{6}, \circled{7}, \circled{8}) and $\Rightarrow$(\circled{1}, \circled{2}, \circled{6}, \circled{7}, \circled{9}, \circled{4}) show how the actions of the \emph{decision system} or the \emph{fraud investigator} impact \emph{fraudulent users}. The fraudulent user may consequently adapt their behavior, i.e., \textbf{adversarial drift}. However, the behavior of \emph{genuine users} can also change, i.e., \textbf{natural drift}. Moreover, several decisions may be taken at different stages in the life-cycle of a business entity, resulting in \textbf{upstream models}. We discuss methods to deal with the adaptivity that this requires from the \emph{ML model} in~\autoref{sec:concept_drift}.}
    \item[ML-investigator interaction]{The cycle $\Rightarrow$(\circled{6}, \circled{7}, \circled{9}, \circled{10}) highlights the ability of fraud analysts to provide labels to aid ML models. One objective is to investigate the most suspicious instances (i.e., \emph{exploitation}). A contrasting objective is to investigate those instances that are expected to be the most informative to the model (i.e., \emph{exploration}). This creates an \textbf{explore/exploit trade-off} regarding which instances are presented to the investigator through \circled{9}. We discuss the aspects involved in the interaction between the ML model and \emph{fraud investigator} in~\autoref{sec:ml_initiated_investigations}.  
    }
    % \item [ML with human experts in the loop]{Labels from \emph{fraud investigators} are used by the \emph{ML model} through \circled{10} $\Rightarrow$ \circled{6}. We discuss how labels can be generated, e.g., through labeling in the closed loop of \circled{6} $\Rightarrow$ \circled{7} $\Rightarrow$ \circled{9} $\Rightarrow$ \circled{10} or by searching through \circled{3} $\Rightarrow$ \circled{10}, in~\autoref{sec:ml_initiated_investigations}}.
    \item[Model]{Relation \circled{6} concerns the training and deployment of the \emph{ML model}. The fraud detection setting has particular requirements for model \textbf{deployment} and \textbf{monitoring}, which we discuss in~\autoref{sec:model_training}.}
\end{description}

\section{Investigation Support}
\label{sec:presentation}

Relations \circled{3}, \circled{9}, and \circled{11} in~\autoref{fig:operational_model} describe the investigations of potential fraud instances that were either found proactively, presented by the \emph{decision system}, or escalated. \emph{Fraud investigations} are often time-consuming and require a high amount of experience and expertise. Much of the time of a fraud investigation goes to gathering evidence and documenting the decision with relevant evidence. 

The number of investigations that can be processed is limited because fraud investigations are time-consuming. Therefore, any support from machine learning in aiding the evidence gathering and decision support is of great benefit. The aim is to make evidence gathering \emph{more efficient} and \emph{more effective}, respectively resulting in the ability to \emph{process more investigations} and to \emph{increase the accuracy of the investigation outcome}.

\subsection{Summary of the Literature}

%\subsubsection{User interfaces} of the investigative tools, and their importance to successful evidence gathering, have barely been studied. Beaugnon et al.~\cite{Beaugnon2018End2End} found that the \emph{graphical user interface} that fraud investigators use for their investigations is an important factor to their performance. Dilla and Raschke~\cite{Dilla2015} analyzed the importance of how the information and evidence is visualized to the fraud investigators and found that the influence of visualization is high especially in complex fraud scenarios.

\subsubsection{Explainable AI} methods and visualizations thereof~\cite{Amershi2014} can provide decision support to the fraud investigator when embedded in the user interface of investigation tools. Weerts et al~\cite{Weerts2019} found no strong evidence that SHAP model explanations increase the \emph{accuracy} and \emph{efficiency} of fraud investigators' decision-making. However, in many fraud detection systems, feature interactions are important to fraud detection accuracy. Therefore, one can hypothesize that rule-based explanations such as \emph{anchors}~\cite{Ribeiro2018AnchorsHM}, which in contrast to SHAP explain the predictions in terms of rules over multiple features rather than in contribution-scores of individual features, might be better suited for the fraud detection setting. More generally, more research is needed into how fraud investigators can be best supported through model explanations.

In the limited research on the use of model explanations for decision support of the fraud investigator, some work (e.g.,~\cite{Bansal2020DoesTW}) uses crowdsourced labelers (e.g., from Amazon Mechanical Turk) for the experiment. The \emph{fraud investigators} in industry are highly trained, therefore, it is questionable whether empirical results that are obtained with untrained crowdsourced participants transfer to a real-life setting with highly-trained fraud professionals.

In contrast to \emph{local interpretability} methods that explain individual predictions, \emph{global interpretability} methods provide insight into how a model as a whole makes its decisions and might be useful to increase the overall trust of fraud investigators and other stakeholders from the anti-fraud department in the ML model. To the best of our knowledge, empirical work on whether global interpretability methods increase model trust in fraud detection settings is lacking.

\subsubsection{Multiple instance learning}
(MIL) allows ML models to classify whole groups of instances at once instead of single instances.
In many fraud use cases, the fraudulent user targets multiple business entities through repeated actions on the e-commerce platform. For example, the same fraudster might perform multiple attempts to compromise accounts. 
%In such cases, in comparison to investigating each instance in isolation, it can help the fraud investigator by providing additional context when a single investigation is performed on all instances related to that same user, i.e., \emph{group-level investigations}. 
In such cases, performing \emph{group-level investigations}, i.e., investigating multiple instances related to that same actor, yields more appropriate evidence and leads to taking action on more instances per fixed unit of time. MIL is the ML-counterpart of group-level investigations, where individual instances are grouped in Bags, e.g., multiple energy states (instance) of a molecule (bag) in drug discovery, or multiple segments (instance) of an image (bag). Surveys of MIL can be found in~\cite{Amores13,CarbonneauCGG16}.

Multiple instance learning methods can aid \emph{group-level investigations} of fraud investigators by identifying groups of instances that could be fraudulent. The bags that are presented to fraud investigators must be relevant, i.e., they must contain a sufficient share of fraud. This is addressed by appropriately defining bag-level labels~\cite{FouldsF10,CarbonneauCGG16}. There is often a trade-off between bag-level and instance-level model performance~\cite{CarbonneauCGG16}. In manual investigations, the former might be more important, while the latter might be more important for automated decisions. 

\subsubsection{Network learning}
is closely related to grouped investigations. Graph-based visualizations can show the fraud investigator which instances are connected based on some identifier, e.g., based on \emph{IP address} or \emph{e-mail address}. Like group-level investigations, the graphs provide the fraud investigator with visual information on which instances are connected to some identifier, and which might therefore possibly also be fraudulent. Network learning~\cite{Liang2019} and graph neural networks~\cite{Hamilton2020} are ML counterparts of graph-based investigations. Such models can aid the fraud investigator by identifying graph nodes or subgraphs where the fraud investigator is likely to find fraud.

%\subsubsection{Instance clustering} can be used to present multiple instances to the \emph{fraud investigator} simultaneously such that a joint decision can quickly be made about the whole cluster. Potentially, this could in some use cases improve the efficiency and accuracy of the investigator's decision-making. This can for example be the case when a non-obvious fraud case is grouped together with another fraud case that is easy-to-detect for the fraud investigator and shares multiple strong indicators of identity. Weerts et al.~\cite{Weerts2019b} apply \emph{case-based reasoning} to select historic instances that are highly similar to a case that a fraud investigator is currently investigating and provide context on the decisions that had been made in these similar cases. This further reinforces the hypothesis that the presence of the second easy-to-detect fraud case can help the fraud investigator to see that the related case is fraudulent as well. However, the clustering can also increase the fraud investigator's decision-making difficulty of the if the instances in the grouped are not informative about each other. 

%Otherwise, \emph{fraud investigator}s can dig deeper and search for fraudulent instances at their own initiative, developing search queries based on the ML model's \emph{feature values}~\cite{Sculley2011AA}, which is an effective supplement to a model-to-investigator feedback loop~\cite{Attenberg2010WhyLW}.

\subsection{Open Research Challenges}
\begin{description}
    %\item[User interface] Since a fraud investigator's work is largely done via a user interface, can the information therein be presented in a way that aids the investigation? Are some classes of visualizations more or less effective at supporting fraud investigations?
    \item[Challenge 1:]{\textbf{Model explanations for decision support}. There is limited research on the effect of model explanations on the \emph{fraud investigator}'s decisions quality and efficiency. It is unclear if these explanations sometimes can bias decisions. It is also unclear what types of model explanations empirically would be most helpful to the fraud investigator, and whether or not this depends on aspects like the application domain, or the experience level of the fraud investigator.}
    %\item[Evidence grouping] Should the fraud investigators be presented with individual instances one-by-one, or with groups of related instances such that a joint decision can be made about the whole group at once?
    %\item[Feature searching] In conventional cases, an investigator performs queries based on aspects of the raw data, such as IP addresses or purchase history. In the same manner that a data scientist would perform feature engineering, can more complex features be derived from raw data that can serve as the basis for investigator's queries, to improve the accuracy and efficiency of their decisions?
    \item[Challenge 2:]{\textbf{Multiple instance learning}. As pointed out in~\cite{CarbonneauCGG16}, most of the current literature on MIL covers applications in biology and chemistry, computer vision, document classification, and web mining. To our knowledge, the area of fraud detection in e-commerce, especially the interaction with fraud investigators, has received little attention in the literature, although related topics like the detection of fraudulent financial statements have been discussed in~\cite{Kotsiantis2008}, as well as HTTP network traffic in~\cite{Pevny2020}. A particularly interesting challenge is how to compose the bags. In practice, this is often done using characteristics of the user actions (e.g., the IP address and date of the user action cf.~\cite{Xiao2015}). We are not aware of MIL literature that addresses bag construction, especially in the context of fraud detection in e-commerce.}
\end{description}

\section{Decision-Making}
\label{sec:decision_making}
Relations \circled{8} and \circled{9} in~\autoref{fig:operational_model} show that the \emph{decision system} can take \emph{automated action} or \emph{trigger an investigation} for the instances that it suspects of fraud. 
The \emph{consequences} of wrong decisions can be severe. False positives respectively lead to adding friction for or blocking a genuine user, or wasting the time of fraud investigators. False negatives result in allowing fraudulent behavior. Another, less severe, wrong decision is to trigger an investigation for a fraud case instead of automatically blocking it. It is the decision system's task to \emph{manage risk} by appropriately trading off risks of different types of wrong decisions, by \emph{combining ML and heuristic rules} that are developed by fraud investigators.
% The former option prevents time investment from the \emph{fraud investigator} and can therefore be a better choice for more obvious fraud cases or when synchronous actions are required. The latter is more careful, as false positives can still be overridden by the investigator. 

Relations \circled{5}, \circled{7} in~\autoref{fig:operational_model} highlight the two types of information sources that the \emph{decision system} has available to make its decisions: \emph{ML models} and \emph{rules}. Models and rules aim to complement each other, and it is the task of the \emph{decision system} to aggregate them into a single action when models and rules might recommend different actions for the same instance.

% \emph{Threshold setting} is the task to map the scores from an ML model to a concrete action. In practice, the \emph{decision system} has more sources of information to base this decision on, and the decision can be based on a combination of rules that are engineered by the \emph{fraud investigators} and possibly even multiple ML models (e.g., a supervised model and an anomaly detection model). Additionally, the design of the \emph{decision system} should take into account the concrete \emph{business objectives} and \emph{metrics} instead of just minimizing the rate of misclassifications.

\subsection{Summary of the Literature}
\subsubsection{Probability calibration} aims to transform the model output of a classifier in such a way that the predicted model score approximately matches the probability of an instance belonging to the positive class. Calibrated model scores play an important role in \emph{managing risk} in decision-making, as it is the prerequisite of some \emph{cost-sensitive learning} techniques (see below). They also enable the calculation of \emph{expected values} of \emph{key performance indicators} of the business. 
%Some classifiers, like logistic regression, natively produce calibrated probabilities. 

Methods for probability calibration include Platt scaling~\cite{Platt1999}, Beta calibration~\cite{Kull2017}, and isotonic regression~\cite{Zadrozny2002}. Such methods require a \emph{calibration set} of data that is held out from the training data. The adversarial concept drift (see~\autoref{sec:concept_drift}) in fraud settings and the sometimes rapidly changing prevalence (i.e., fraud rate) bring difficulties in obtaining model scores that are close to probabilities \emph{in the latest production data} and not just on the calibration set.

\subsubsection{Cost-sensitive learning}
takes the misclassification costs into account and thereby enables making decisions that minimize the expected cost of fraud to the business operations, rather than simply minimizing the number of classification errors~\cite{Ling2010}. This addresses the challenge of \emph{managing risk} in decision-making. The application of cost-sensitive learning requires the estimation of the \emph{costs} (or benefit) to the business of the four cells of the confusion matrix: the \emph{false positive}, \emph{false negative}, \emph{true positive}, and \emph{true negative}. The cost of a \emph{false positive} could for example be the missed income from a blocked transaction, while the cost of a \emph{false negative} could be the financial cost of that fraud instance. 

Practical difficulties often arise because some aspects of these costs can be \emph{difficult to quantify or measure}, such as the \emph{reputational damage} to the business in the case of misclassification. More research is needed on guidelines and frameworks for \emph{how} to design cost functions for cost-sensitive learning when some aspects of the costs are not easily quantified financially.

Once cost functions are in place, the expected costs can be calculated trivially if the classifier returns calibrated scores; alternatively, a method like \emph{empirical thresholding}~\cite{Sheng2006} can be used to minimize the expected costs under uncalibrated model scores. The label might not be available in cases where a fraud attempt was blocked by an automated action. In that case, one might make use of the control group or some other source of unbiased data (see~\autoref{sec:feedback}) to optimize the threshold with regard to the cost function.

\subsubsection{AI fairness} is an important topic in fraud detection because misclassification can have \emph{severe consequences} -- a false positive often harms genuine users, like purchases being canceled or accounts being disabled. Since any decision taken in fraud detection potentially directly effects a real person, attention should be paid to ensure fairness and mitigate fairness issues where they may exist.

The body of literature covering fairness in ML is extensive~\cite{Mehrabi2019ASO}, including considerations for its practical applications in production systems~\cite{Beutel2019PuttingFP,Holstein2019ImprovingFI}. While not much fraud-specific applications research has been done in ML fairness, the problem can be cast more generally as a supervised classification setting where positive predictions signify actions taken against individuals. Similar aspects can be found in mortgage default prediction~\cite{Hardt2016EqualityOO} or recidivism prediction~\cite{Chouldechova2018ACS}. One particular challenge to e-commerce organizations looking to achieve ML fairness is \emph{low observability} on certain protected attributes -- dimensions such as race or sex are often not explicitly collected in e-commerce platforms, rendering some attribute-dependent ML fairness methods~\cite{pmlr-v97-ustun19a} inapplicable.

The presence of an adversary (fraudster) in the fraud context presents a unique challenge to the implementation of fairness controls. Protected attributes can be spoofed by dishonest actors, for example, by using a VPN to pretend to be in a different country. If the system uses a fairness control that applies corrections conditioned on these protected attributes, the fraudster may be able to tweak their attributes to maximize their success. Data poisoning attacks that target fairness controls have been recently developed~\cite{Solans2020PoisoningAO,Mehrabi2020ExacerbatingAB}.

Anomaly detection is also important to fraud detection for flagging novel and potentially malicious behaviors but has its own set of fairness pitfalls. Recent work ~\cite{Davidson2020AFF,Shekhar2020FAIRODFO} aims to quantify and mitigate these issues, but overall, fairness in anomaly detection systems is still a novel area of research.

\subsubsection{Uncertainty quantification}
has clear applications in the settings of \emph{active learning} (see~\autoref{sec:ml_initiated_investigations}) and \emph{classification with a reject option}~\cite{Chow1970,Hellman1970}, where the ML model has the option to refuse or delay making a decision when uncertainty is too high around the model's prediction. H{\"u}llermeier and Waegeman~\cite{Hullermeier2021} provide a detailed survey of methods for \emph{uncertainty quantification}. 
 
% A naive ML solution would be to simply add users that a fraud detection ML model scores below a certain threshold. However, for new attack patterns that might arise, it is uncertain how an ML model would score them, and fraudulent users should be prevented from reaching the list of trustworthy users. .%

\emph{Classification with a reject option} has applications in fraud detection for so-called \emph{trust systems}, which are tasked with assigning a permanent trust status to some subsets of the \emph{genuine users} that are important customers and are clearly not fraudulent users, for which taking any action on them should at all times be avoided. Accidentally trusting fraudulent users could do a lot of damage, and therefore making a prediction can be rejected by the model when there is too much uncertainty regarding the instance. %For example, a webshop like Amazon might want to exempt their largest merchants from \emph{automated actions} \circled{6}. One approach to achieve this is to maintain a manually curated list of so-called \emph{trustworthy users} and to prevent the \emph{decision system} from taking action on those. % Alternatively, hand-crafted rules \circled{5} can be developed to identify trustworthy users, or machine learning could be used.

An important concept in both \emph{classification with a reject option} and \emph{active learning} is \emph{epistemic uncertainty}, i.e., the degree of uncertainty due to lack of data in the part of feature space for which the prediction needs to be made (reducible uncertainty). This contrasts \emph{aleatoric uncertainty}, i.e., the degree of uncertainty due to an overlap in class distributions in the part of input space for which the prediction needs to be made (irreducible uncertainty). Methods that explicitly quantify the \emph{epistemic} part of the uncertainty and can separate this from the \emph{aleatoric} part are summarized in~\cite{Hullermeier2021} and include \emph{density estimation}, \emph{anomaly detection}, \emph{Bayesian models}, or the framework of \emph{reliable classification}~\cite{Senge2014}. The argument is that rejecting or delaying a decision is only a reasonable decision if it is expected that the uncertainty is expected to \emph{decrease}, which is not the case with aleatoric (irreducible) uncertainty. Likewise, in the active learning setting, spending the fraud investigator's time to investigate an instance only makes sense if there is reducible uncertainty regarding that instance. Marking a user as trustworthy seems safe when a model predicts a user to be of the non-fraud class with \emph{low epistemic uncertainty} regarding that prediction. However, quantification of \emph{epistemic uncertainty} is a rather novel research direction in the field of machine learning, and more research is needed. More specifically, the application of epistemic uncertainty quantification in adversarial problem domains is not well-understood.

\subsubsection{Rules-based systems}
are created by \emph{fraud investigators} and are designed to supplement ML models in the detection of fraud. Because of concept drift, rules can become ineffective shortly after they have been added to the \emph{decision system}. After fraud investigators have first observed a new type of fraud attack that is not recognized by the ML model, a rule can be used for the period until the ML model picks up on that attack. In some sense, decision-making based on model output and multiple rule outputs can be seen as analogous to \emph{ensemble learning}~\cite{Sagi2018}, which concerns the decision-making based on multiple ML models. A specific challenge is that the set of rules is subject to change: new rules get developed and old ones get decommissioned. This creates a need for ensemble models that combine a non-stationary set of components.

\subsection{Open Research Challenges}
\begin{description}
    \item[Challenge 3:]{\textbf{Model calibration under adversarial drift}. Existing model calibration techniques help to generate calibrated probabilities on instances that are identically distributed as the calibration set. There are open questions whether there are any bounds to the degree to which calibration can break down in an adversarial drift setting and whether this can be mitigated.}
    \item[Challenge 4:]{\textbf{Guidelines for design of cost-sensitive learning cost functions}. There is a need for frameworks and guidelines for the engineering of cost functions for cost-sensitive learning applications in situations where not all relevant factors of the cost are easily expressed in financial value.}
    \item[Challenge 5:]{\textbf{Fairness and vulnerability}. There are open questions regarding whether AI fairness methods could open up potential vulnerabilities in adversarial settings. Furthermore, there are open challenges around fairness in scenarios when the sensitive attributes can be spoofed by fraudsters.}
    \item[Challenge 6:]{\textbf{Fairness in anomaly detection}. AI fairness in the area of anomaly detection systems is still a novel area of research. It is unclear if post-processing methods for AI fairness that work in the supervised setting are applicable in the setting of anomaly detection.}
    \item[Challenge 7:]{\textbf{Epistemic uncertainty quantification for trust systems}. Predictions of the non-fraud class that are made with \emph{low epistemic uncertainty} could have an application to  identify trustworthy users that should never be marked as fraudulent. However, more research is needed into the applications of such techniques in adversarial problem settings.}
    \item[Challenge 8:]{\textbf{Ensemble learning for non-stationary sets of components}. The output of the ML model and the rules ultimately need to be combined into a single decision. Ensemble learning methods address this task but do currently not handle dynamic sets of ensemble components.}
\end{description}

\section{Selection Bias in Labels}
\label{sec:feedback}
% Feedback collection is the process of collecting labels (e.g., \emph{fraud} or \emph{non-fraud}) for instances for training or evaluating fraud detection systems. From the \emph{machine learning} perspective, we would like to have a random uniform sample of labeled transactions, since inference is done over all the transactions in the platform. There are two main sources of labeled examples: \emph{fraud investigators} \circled{10}, and \emph{automatic escalations} \circled{12}, each presenting its own challenges.

% %\subsection{Automatic Escalation Feedback}
% Some fraud use cases (partially) obtain labels through \emph{automatic action}s and \emph{automated escalation}, i.e. the cycle $\Rightarrow$(\circled{1}, \circled{2}, \circled{6}, \circled{7}, \circled{8}) in~\autoref{fig:operational_model}. This introduces the following challenge: once a transaction is rejected, its true label won’t be available -- on the other hand, if a transaction is approved, its true label might be available after some significant amount of time. This implies that the labeled data is always biased towards older and approved transactions. 

Labels are obtained through \emph{fraud investigations} \circled{10}, and \emph{automatic escalations} \circled{12}. From the \emph{machine learning} perspective, we would like to train models using labeled instances that are uniformly sampled from the population. In practice, there are several sources of \emph{selection bias}. First, \emph{delay in labeling} arises for both label sources: manual investigations can take minutes up to days, whereas it can take days up to weeks for notifications of fraud to arrive through escalations~\cite{DalPozzolo2015}. Secondly, manual investigations may \emph{overlook} fraudulent instances (e.g., caused by resource constraints, or well-hidden fraud). Third, \emph{automated actions} \circled{8} block suspicious transactions, and as a result, there will be merely a suspicion (no documented evidence) that these transactions are fraudulent. Because there is no certainty that these blocked transactions are fraudulent, they cannot be considered to be labeled instances. In many e-commerce applications, similar issues are commonly addressed with a \emph{control group}, i.e., by always approving a certain percentage of transactions. While this control group would be an unbiased sample of labeled data, collecting it would come at the high cost of needing to purposefully let a share of fraud go through without blocking it. %Paradoxically, the value of this control group is proportional to the amount of fraud in the control group. 

\subsection{Summary of the Literature}

\subsubsection{Learning under selection bias}
has been studied for example in~\cite{Zadrozny2004} and~\cite{swaminathan15}, both relying on ideas from \emph{causal inference} such as \emph{inverse propensity weighting}. Another example worth mentioning is~\cite{smola2007} where \emph{matching} weights are computed directly. In~\cite{jacobusse2016}, the authors study the problem in the low prevalence regime, which is particularly fitting the fraud detection problem, although they don't construct an unbiased model. Rather, they propose to utilize unlabeled data to construct a case ranking model, which might or might not be appropriate depending on the specific problem at hand. The \emph{domain adaptation} field also tackles this problem, defined as learning a model with data sampled from a \emph{source domain} to be applied in a \emph{target domain}. This particular field distinguishes several settings, two of which are of particular interest to fraud detection: 
\begin{description}
    \item[Unsupervised domain adaptation] (e.g.,~\cite{ganin15,Sun2016}) considers labeled and unlabeled examples from the source domain and unlabeled examples from the target domain, matching the no-control-group setting.
    \item[Semi-supervised domain adaptation] (e.g.,~\cite{daume2010frustratingly}) adds some labeled examples from the target domain, matching the with-control-group setting.
\end{description}

\subsubsection{Multi-armed bandits}
(MAB)~\cite{Lattimore2020} is an area of research that studies the trade-off between \emph{exploration} and \emph{exploitation}. A control group is a simple form of exploration. In the context of automated actions, approving a transaction allows us to observe both the consequences of approving as well as the consequences that would have been observed if the transaction would have been blocked. This is known as \emph{partial feedback} and is different from the \emph{bandit feedback} setting where feedback is only observed for the action that was taken. No feedback is observed for blocked transactions.

To handle this setting, one can create more sophisticated exploration policies, that don't necessarily approve cases uniformly at random but explore with some optimization criteria. The principle of \emph{optimism in the face of uncertainty}~\cite{auer2002using,li2010contextual,mahajan2012logucb} is of particular interest for the fraud detection problem. The core idea is to prioritize exploration (acceptance) of transactions where the expected cost is lower or the model has higher uncertainty. Typically, the expected cost is estimated through standard supervised learning techniques and the uncertainty is modeled with the variance of the mean cost estimate. Other approaches such as \emph{Thompson sampling}~\cite{thompson,agrawal2012analysis}, or more generally, \emph{posterior sampling}~\cite{russo2014learning} can balance exploration and exploitation in fraud detection.

\subsection{Open Research Challenges}
\begin{description}
    \item[Challenge 9:]{\textbf{Bias-variance trade-off}. Removing the bias from the data almost always involves an increase in the variance of the predictions. This variance might lead to models with poor generalization error, defeating the purpose of bias reduction. Most bias reduction techniques focus on completely removing the bias, and although there exists work on variance reduction, it is always under the no-bias constraint. Creating principled mechanisms to tune this trade-off, potentially allowing positive bias but improving generalization error is still an open challenge, and an active area of research, mainly in the \emph{domain adaptation} field. A related and harder challenge is the fact that in practice, at training time, there is no data available from the \emph{target domain}, which can be considered an adversarial version of the \emph{unsupervised domain adaptation} problem where the goal is to learn a model that generalizes \emph{sufficiently well} to a large set of potential target domains from labeled source domain data.}
    \item[Challenge 10:]{\textbf{Pseudo-MAB setting}. Only the approval action reveals full feedback whereas rejection reveals no feedback. This setting does not exactly match the MAB setting. This opens questions about the optimality of standard MAB policies. An alternative formulation is simply to select a subset of transactions for rejection (or acceptance) to minimize some carefully crafted loss function that combines the monetary costs with the value of the gathered information. This can be addressed from the perspective of set-function optimization and \emph{online active learning}~\cite{sculley2007online}. However, the MAB formulation addresses other relevant challenges such as delayed feedback and non stationary which have been studied to a large extent in the MAB literature (e.g., \cite{joulani2013online,gyorgy2020adapting}).}
\end{description}

\section{Concept Drift}
\label{sec:concept_drift}

In~\autoref{fig:operational_model}, the cycles $\Rightarrow$(\circled{1}, \circled{2}, \circled{6}, \circled{7}, \circled{8}) and $\Rightarrow$(\circled{1}, \circled{2}, \circled{6}, \circled{7}, \circled{9}, \circled{4}) highlight how fraud detection is an adversarial problem domain. When the \emph{decision system} is successful in blocking the fraud attempts of a fraudulent user (i.e., \circled{4} or \circled{8}), then the fraudster is likely to try to circumvent the system by modifying their attack until successful. Due to this behavior, fraud detection systems often experience concept drift nearly constantly.

Besides adversarial drift from changing fraud attacks, the data distributions that are generated by genuine users can also be subject to concept drift. Examples include seasonal patterns, unexpected events (e.g., COVID-19), or changes in the e-commerce platform. However, the drift of genuine users is often largely independent of \circled{4} and \circled{8} and is thus not adversarial.

The third source of concept drift to fraud detection systems is related to updates to so-called \emph{upstream} models. For example, imagine a webshop that requires users to log in before they can make a purchase. An update to a login-time fraud detection model shifts the distributions of the data that reaches a payment-time fraud detection model that occurs later in the sales funnel, because the population of fraudulent users that are already caught by the login-time model will likely change with the update.

\subsection{Summary of the Literature}
\subsubsection{Concept drift adaptation}~\cite{Gama2014,Lu2018}, or \emph{dataset shift}~\cite{Quinonero2009}, is a well-studied research topic. Drift can be categorized by their \textbf{distributional type}: \emph{covariate shift} concerns a shift in $P(X)$, \emph{prior shift} a shift in $P(y)$, and \emph{real concept drift} a shift in $P(y\mid X)$. Orthogonally, drift can be categorized by its \textbf{temporal type}: it can be \emph{sudden}, \emph{gradual}, \emph{incremental}, or \emph{recurring}. Finally, drift can be \emph{adversarial} or \emph{natural}, i.e.,   \emph{adversarial} drift is specifically aimed to beat a detection system, while  \emph{natural} drift happens for reasons that are exogenous to it.\looseness=-1

Fraud detection has adversarial drift in the \emph{fraudulent} class. Changes to the attack patterns of fraudsters often result in gradual and incremental drift, as fraudsters tend to gradually increase the frequency of their successful and undetected attacks while decreasing the frequency of detected and unsuccessful attacks. Fraudsters can also cause recurring drift, as they occasionally retry old attempts to check whether the fraud detection systems still catch them.

%Changes to an upstream model can substantially change the distributions for a model that is situated downstream. For example, the upstream model might detect fraud patterns that were previously not detected resulting in the instances being no longer present in the distributions for the downstream model. Note that such drift is introduced suddenly, i.e., with the deployment of the upstream model.

Drift due to changes in the e-commerce platform is often \emph{sudden} because changes (e.g., in an account registration portal or a payment process) change at once at the time of new code deployment. However, in practice, many changes in the e-commerce platform are first evaluated in an A/B test, and thus, the drift that results from this change might at first affect only a fraction of the users. Furthermore, the drift that results from changes in the e-commerce platform is natural drift, contrasting the adversarial drift that originates from fraudsters' attempts to remain undetected. Much of the adversarial concept drift detection and adaptation literature ignores that such tasks often need to be performed \emph{in the presence of sudden and natural drift} that originates from changes to the platform itself. While many concept drift detection techniques exist~\cite{Gama2014}, there is a practical need for methods that can distinguish the fraudsters' gradual adversarial drift from the sudden and natural drift that is caused by platform changes.\looseness=-1

\emph{Delayed labels} make the task of concept drift adaptation much more challenging. Until the labels are known, concept drift is only detectable when a change in $P(y\mid X)$ is accompanied by a change in $P(X)$~\cite{Zliobaite2010}. Likewise, adaptation to a change in $P(y\mid X)$ is not possible without a change in $P(X)$. Several methods exist to address the problem of concept drift adaptation under delayed labels, including \emph{positive unlabeled} (PU) learning~\cite{Kiryo2017,charles2008}, or by explicit modeling of the expected label delay of individual instances through survival modeling. Dal Pozzolo et al.\cite{DalPozzolo2015} proposed a solution specific for the fraud detection case where they train two separate models. The first model is trained on the labels found by \emph{fraud investigators}, while the second is trained on labels obtained through the often much more delayed label-source of escalations. In practice, some fraud detection use cases deal with label delay that is theoretically upper bounded, such as in the case of credit card chargebacks that have a deadline set by the credit card issuers. To the best of our knowledge, concept drift detection under \emph{upper-bounded label delay} has not been studied as of yet.

\emph{Supervised methods} for fraud detection often outperform purely unsupervised anomaly detection for fraud detection in industry applications~\cite{Gornitz2013}. Supervised models, in particular, outperform anomaly detection models in detecting fraud instances that are continuations of fraud attacks that were ongoing at the time the model was trained. The field of \emph{evolving data stream classification} developed several methods to incrementally update ML models in a streaming setting to adapt to distributional changes. State-of-the-art methods include \emph{adaptive random forest}~\cite{Gomes2017} and \emph{streaming random patches}~\cite{Gomes2019}. The field is heavily focused on \emph{updating} ML models instead of retraining them from scratch, which is motivated by computational efficiency. In practice, however, e-commerce organizations do have the computational resources that are required to retrain models daily. %However, periodic retraining of ML models does comes with several open challenges, which we discuss in~\autoref{sec:model_training}.

Finally, there is currently limited insight into \emph{how} fraudsters respond, adapt their attacks, and cause drift. In practice, the fraudsters don't have \emph{direct} control over the feature vectors that their attacks produce, but they instead control it only \emph{indirectly} through their interactions with the e-commerce platform. This constrains how fraudsters can change the distributions of feature values that they generate. The field currently lacks methods to identify potential fraud attacks that the e-commerce platform theoretically would allow for but that have not yet been observed. One possible direction is the use of \emph{attack trees}~\cite{Mauw2005}, a common method in the cybersecurity field to map out possible attack angles for hackers. 
%Open challenges are \emph{how to map the possible attack angles for fraud in an e-commerce platform}, and \emph{how to leverage it in strategic decision-making and prioritization in ML systems for fraud detection}?

\subsubsection{Adversarial robustness in ML}~\cite{Biggio2018} is a research area that focuses on building ML models that make it hard for attackers to create \emph{adversarial examples}, i.e., data points that the model predicts wrongly. Adversarial robustness closely links to concept drift--a fraud detection system that is adversarially robust makes it more difficult for fraudsters to generate new types of fraud that remain undetected. Current work on adversarial robustness is heavily focused on computer vision and natural language processing tasks, while the majority of fraud detection systems use tabular data. Adversarial robustness methods for tabular data are an open research challenge with applicability in fraud detection.

\subsubsection{Anomaly detection}
is a class of methods that separate \emph{normal} data points from \emph{outlier} data points. The task of anomaly detection strongly links to \emph{density estimation}, and can be seen as its inverse. \emph{Novelty detection}~\cite{Pimentel2014} concerns the detection of \emph{novel} behavior that emerges after drift and is therefore of particular relevance to fraud detection. Novelty detection typically uses anomaly detection--what is normal w.r.t. pre-drift data is likely to be an outlier w.r.t. post-drift data. Several empirical benchmark studies~\cite{Aggarwal2017,Emmott2015} have compared anomaly detection methods, often identifying \emph{isolation forest}~\cite{Liu2008} performs well consistently.

New attacks by fraudsters generate feature values that are distinct from their previous attacks, causing the new attacks to be marked as outliers. However, a drift in the behavior of genuine users (e.g., due to changes in the e-commerce platform or exceptional events like COVID-19) is also likely to generate feature values that are distinct from earlier behavior. Therefore, not every outlier can be assumed to be fraudulent, and not every distribution shift is caused by a change in fraud attacks. Marking all outliers as possible fraud cases that require investigation by the \emph{fraud investigator} introduces spikes of false positives. In practice, such a spike would for example be expected with every release of a new change in the e-commerce platform. This calls for investigation into methods that account for the existence of ``harmless" outliers caused by external changes.

\subsection{Open Research Challenges}
\begin{description}
    \item[Challenge 11:]{\textbf{Separating platform changes from changes in attack patterns}. Changes in the e-commerce platform and changes in fraudster behavior both drive concept drift. In the former case, drift tends to be natural and sudden, while in the latter case it tends to be adversarial and gradual. Concept drift detection methods that alert in the latter case but not in the former would be of practical value for fraud detection.}
    \item[Challenge 12:]{\textbf{Accounting for platform changes in novelty detection}. Changes in the e-commerce platform can cause large spikes in the number of outliers that are flagged by novelty detection algorithms, thereby limiting their practical use. This creates a need for methods that aim to detect outliers that are novel fraud types, but not outliers that result from platform changes.}
    \item[Challenge 13:]{\textbf{Mapping attack angles}. There is a need for methods and frameworks to map possible attack-angles in an e-commerce platform, and for decision-making frameworks to leverage those this in work prioritization.}
    \item[Challenge 14:]{\textbf{Methods for adversarial robustness for tabular data}. Many fraud detection systems work on tabular data, which is an understudied data modality in the research field of adversarial robustness.}
    \item[Challenge 15:]{\textbf{Balancing anomaly detection and supervised methods}. While \emph{supervised methods} are more accurate in detecting recurring fraud types, \emph{anomaly detection methods} can detect new attacks. Can a strategy for combining both types of models be automatically inferred?}
    \item[Challenge 16:]{\textbf{Concept drift adaptation in the delayed label setting}. \emph{Supervised methods} for \emph{concept drift adaptation} often assume that labels are immediately available. How do we adapt to concept drift if labels may be delayed? In some fraud problems, there is a theoretical upper bound in the label delay. Can this upper bound be used in concept drift adaptation?}
\end{description}

\section{ML-investigator interaction}
\label{sec:ml_initiated_investigations}
Fraud investigations are a vital part of the operational model by stopping fraudulent behavior through manual action \circled{4}, and as a result, generating new labels for the ML model \circled{10}. Two objectives are involved here: the goal to \emph{identify fraud} and the goal to \emph{generate labels that are most useful to the model}. These two objectives can sometimes compete. Below we describe various machine learning (ML) methods to trigger investigations \circled{9} in ways that address and balance these objectives, and we discuss open research questions.

% not sure if this reference is useful somewhere: \cite{Sculley2011}

\subsection{Summary of the Literature}
\subsubsection{Active learning}
(AL) is a paradigm that naturally fits cycle $\Rightarrow$(\circled{6}, \circled{7}, \circled{9}, \circled{10}) in~\autoref{fig:operational_model}. The AL paradigm decides which unlabeled data points to prioritize for labeling depending on how much they are expected to improve the ML model. \emph{Fraud investigators} label these data points, after which the model can be retrained and a new iteration of data point prioritization is started. AL can help to generate rapidly when escalations (i.e., \circled{11} and \circled{12}) are slow. This helps to mitigate label delay (see~\autoref{sec:feedback}) and therefore improves the model's ability to adapt to concept drift (see~\autoref{sec:concept_drift}). Furthermore, selecting instances that would maximize the learning is an efficient use of the fraud investigator's time.

AL has not been widely applied in industry context so far despite extensive academic research~\cite{attenberg2010,settles11a}. This might be due to uncertainty about which AL technique to use, how to deal with extreme class imbalance, the possibility of viable alternatives, the engineering overhead, and uncertainty about the validity of the assumptions made in AL. The problem of class imbalance is particularly relevant for many fraud detection use cases. Carcillo et al.~\cite{carcillo2018} investigated AL methods under the high class-imbalance setting of fraud detection and showed on credit card fraud data that simply selecting the instances with the highest probability of being fraudulent maximizes learning and obtained high precision. The popular uncertainty sampling~\cite{ALSurveyBurr} method explores those data points with high proximity to the model's decision boundary. More recent work on active learning~\cite{Nguyen2019} aims to distinguish \emph{epistemic uncertainty} from \emph{aleatoric uncertainty} (see~\autoref{sec:decision_making}). The rationale is that the fraud investigator's investigation time is wasted time if they investigate data points in parts of the feature space where the uncertainty cannot be reduced (i.e., where uncertainty is aleatoric). \emph{Inter-rater disagreement} between multiple fraud investigators about the same instance can cause aleatoric uncertainty. Traditional AL methods that do not distinguish between the two uncertainty types tend to repeatedly select instances with high \emph{aleatoric} uncertainty~\cite{Nguyen2019}, thereby wasting time of the fraud investigator. New fraud patterns are likely to come from regions of the feature space with high \emph{epistemic} uncertainty. Therefore, fraud detection AL systems would benefit from focusing on sampling instances based on epistemic uncertainty. This research area is novel and lacks real-life evaluations in fraud settings.

\subsubsection{Guided learning}~\cite{Attenberg2010WhyLW} contrasts by asking fraud investigators to search themselves for fraudulent examples (i.e., $\Rightarrow$(\circled{3},  \circled{10})), instead of being asked to provide labels for specific instances that were selected by the AL model. Guided learning is possible when investigators have sufficient domain knowledge to find positive examples themselves. This is typically the case in the fraud domain. Guided learning can be particularly useful when prevalence is very low and in situations with disjunct classes, like when there are different fraud modus operandi. A disadvantage is that the cost per label for guided learning is most likely higher than for AL. A further disadvantage is that relying on investigator searches induces selection bias that is unique to each investigator, the impact of which has not been studied to the best of our knowledge. Practically, guided learning can be supported by empowering investigators to generate queries based on ML models' input features~\cite{Sculley2011AA}. This approach allows investigators to directly investigate specific areas of the feature space. While it is not a well-studied methodology, this presents a potential area of research.

Guided learning and AL can complement each other: fraud investigators can both proactively search for fraud and label suggestions from an AL model. Some success has been obtained with hybrid variants that start with guided learning and evolve to AL when some initial data set has been gathered ~\cite{Attenberg2010WhyLW}, or that supplement AL with additional searched labels~\cite{Beygelzimer2016}. Guided learning is particularly successful compared to AL in settings with low prevalence~\cite{Attenberg2010WhyLW}. However, the exact success factors in applications of searching and labeling are not well-understood beyond the dependence on prevalence.

\subsubsection{Weak supervision}
techniques such as \emph{snorkel}~\cite{Ratner2020} solve the problem of inferring labels for instances using so-called \emph{labeling functions} that \emph{fraud investigators} create. These \emph{labeling functions} are expected to be imperfect (i.e., \emph{weak}), and can be seen as analogous to the \emph{rules} in relation \circled{5}. The main idea of weak supervision is to infer \emph{reliable labels} from a collection of weak labels using a generative model, which can then be used as ground truth to train the fraud detection model. An important aspect of weak supervision tools is a user interface that allows a \emph{fraud investigator} and an ML practitioner to collaborate and develop new labeling functions that assign labels to instances that are not yet labeled by existing labeling functions. Fraud investigators must be able to quickly find patterns in currently unlabeled data points and develop new rules to apply weak supervision successfully. This procedure is represented by the cycle $\Rightarrow$(\circled{3}, \circled{10}) and by \circled{5}. In existing literature, like~\cite{Ratner2020}, the focus of this user interface is on textual data, where it is for example easy for a \emph{fraud investigator} to instantly spot whether a tweet is \emph{spam} or not. In practice, many fraud detection problems concern tabular data, and much more expert knowledge and deeper investigations are needed for the fraud investigator to conclude if a certain instance is fraudulent. This requires further research for user interfaces that support the iterative process of developing labeling functions for weak-supervision in the context of tabular data.\looseness=-1

%\subsubsection{Human-in-the-loop ML} methods focus on guiding iterative improvements in ML workflows by incorporating domain knowledge from experts in the loop. Helix~\cite{Xin2018} attempts to make steps towards expert-in-the-loop ML by providing a \emph{domain-specific language} to define a full ML workflow (i.e., including feature definitions, data pre-processing, and model training), where the workflow can be visualized in a graphical user interface that allows for discussions.

\subsection{Open Research Challenges}
\begin{description}
    \item[Challenge 17:]{\textbf{Exploration/exploitation trade-off in active learning}. While in typical use cases of active learning the goal is to label data points that are most helpful for improving the model, in the fraud detection use case it is important to trade-off this goal with finding more fraud cases. The trade-off between these two goals is currently an open research challenge.}
    \item[Challenge 18:]{\textbf{Epistemic vs. aleatoric uncertainty sampling}. How can we leverage active learning methods, while avoiding wasting our investigative resources on parts of  the feature space with high aleatoric uncertainty? Epistemic uncertainty sampling is a promising direction of research, but applications of epistemic uncertainty estimates for active learning  in a practical fraud prevention setting with adversarial drift are lacking.}
    \item[Challenge 19:]{\textbf{Label vs. search}. The relative value of AL-type labeling and guided-learning-type searching depends on the cost of the two types of investigations and the fraud prevalence. In many situations, searching is most likely more expensive than labeling, but the exact conditions that influence the costs of both approaches are not clear.}
    \item[Challenge 20:]{\textbf{Weak supervision tools for fraud detection}. The applicability of weak supervision methods is highly dependent on the ability to quickly and accurately assess the class of observed data points. This is often difficult because fraud investigations can be complex and time-consuming. Therefore, better decision support tools for fraud investigators are needed not only to assist the fraud investigations themselves but are also a requirement for applications of weak supervision methods for fraud detection.}
\end{description}

\section{Model Deployment \& Monitoring}
\label{sec:model_training}
Training, deployment, and monitoring of ML models~\circled{6} in a production environment comes with a variety of challenges, some of which are specific to the setting of fraud detection. %such as preventing \emph{technical debt} \cite{sculley2015hidden} and avoiding train-serve skew \cite{tata2017quick,Breck2017}. These challenges are extensively covered in literature. 
Fraud detection models are often integrated into vital parts of the e-commerce platform, such as the payment portal or account registration portal. The financial business consequences are large when such systems malfunction. For example, the business revenue would almost come to a complete halt if an  outage would cause the platform to be unable to process payments or process login requests. Note that this situation is distinct from, for example, a recommender system, where an outage would be undesirable but of smaller consequences. Therefore, it is important to take risk mitigation measures to ensure that model deployment is safe. Additionally, fraud models are particularly retrained and deployed frequently compared to ML models in other parts of the business, because the adversarial drift creates a need to do so (see~\autoref{sec:concept_drift}). This creates a need for deployment safety measures to be efficient and automated.

%The importance of determining the \emph{prediction quality} of a model can depend on the actions that the \emph{decision system} can take. When the decision system solely triggers investigations (i.e., it does not take any automated action), then an accidental increase in false positives could temporarily increase fraud investigator workload, which is straightforward to revert. However, if a model for example directly blocks credit card payments, false positives could result in significant loss of revenue and customer satisfaction. A common technique to estimate the prediction quality is to hold out data from training for offline evaluation. 

% The adversarial nature and commonly unbalanced datasets in the fraud domain highlight some more prominent and unique challenge.

% \subsection{Model Training}
% % Optimalisation
% Fraud detection is frequently mentioned applications in papers about unbalanced classification \cite{dal2015calibrating,ganganwar2012overview}. While fraud does not have to be imbalanced by definition, a healthy business is expected to have more genuine than fraudulent users. 

% -- Class imbalance can be solved by re-sampling data, cost-sensitive learning\cite{elkan2001foundations,ling2008cost}
% -- 

% -- Threshold setting
%  \cite{sheng2006thresholding}
% % Evaluation

% - Offline model evaluation is challenging, as actioning could trigger fraudsters to change behavior.    
% - Search reference of temporaral difference between train/test

\subsection{Summary of the Literature}
\subsubsection{Model verification} methods allow to validate that the ML model satisfies certain desired properties. \emph{CheckList}~\cite{Ribeiro2020} is a verification method inspired by \emph{metamorphic testing} in software engineering and requires the ML practitioner to formulate a set of unit-test-like checks that the model needs to pass. While the main use case of CheckList is during offline evaluation, these unit tests can additionally be used as a sanity check to verify that a model that has been deployed in the model serving platform indeed still passes the unit tests. For natural language data, CheckList provides a mechanism to automatically generate test cases at scale. For other forms of data, formulating test cases is currently still a manual process. Automated test case generation for data formats other than natural language is still an open challenge.

\subsubsection{Deployment best practices} have recently increasingly become a subject of study, and provide guidance on how to manage and mitigate the risks that are involved in model deployment. Early work includes the \emph{ML test score}~\cite{Breck2017}, which provides a list of checks to be performed during model deployment that can catch common problems and mistakes. More recent work includes~\cite{Paleyes2020,Lavin2021}. Commonly recommended is the practice of \emph{canary testing}, i.e., to expose a newly deployed model first to a small group of users where it is closely monitored before exposing all users to the model. Alternatively, in a \emph{shadow mode} deployment, the new model starts making predictions for every instance without being used by the \emph{decision system}. Another common recommendation is to create the infrastructure that allows for quick and safe rollbacks to an earlier model version, which enables quick recovery in case of unforeseen problems. 

\subsubsection{Automated data validation} methods focus on monitoring and validation of the feature values that are used in the ML model. Fraud detection models often consume data generated by parts of the \emph{e-commerce platform} that are often not directly maintained by the anti-fraud department. For example, a payment processing service (or an accounts registration portal) often has a dedicated team that builds the service, owns its database tables, and controls its data schema. These database tables are then read by the fraud detection system to calculate feature values. There is a risk that newly deployed changes (or bugs) in these upstream dependencies affect the feature values and therefore the performance of the production model. For example, the accounts registration portal could contain an \texttt{age} field. A fraud detection model that uses this value as feature is negatively impacted when the team managing the registration portal changes the semantics of the \texttt{age} field by changing it from a mandatory to an optional field.\looseness=-1

Automated data validation for ML applications has been studied extensively~\cite{Polyzotis2019,schelter2018automating,Caveness2020}. Solutions typically perform simple checks, such as validating that all feature values are within a reasonable range of values (e.g., \texttt{age} must not be negative), or by validating that the feature values have a reasonable distribution (e.g., values are not constant). Another frequent approach is to validate that \emph{recent values} of features are within some threshold of similarity compared to \emph{older values} of that same feature (e.g., using Kullback-Leibler divergence). While comparing feature distributions over time is useful to detect \emph{change}, in practice, there is an important distinction between a change to \emph{the semantics of a feature}\footnote{Such as the \texttt{age} field that becomes optional.  Or, alternatively, a \texttt{temperature} field that changes its implementation from Celsius to Fahrenheit.} and \emph{a change in user behavior}. The former case might require the data team to repair data pipelines, while the latter case requires the model to adapt to concept drift (see~\autoref{sec:concept_drift}). There is an open challenge to detect changes in the semantics of features without false alarms on changes in user behavior.

\subsection{Open Research Challenges}
\begin{description}
    %\item[Evaluation loop trade-offs]{How do we select data to set aside for the holdout set to be able to evaluate model reliably, while at the same time having recent data available to be able to adapt to concept drift?}
    \item[Challenge 21:]{\textbf{Test case generation for model verification}. There are existing methods for automated test case generation in the natural language domain. This is an open challenge for other types of data.}
    \item[Challenge 22:]{\textbf{Automated data validation under concept drift}. Existing methods for automated data validation often compare recent feature values to older values. There is a need for automated data validation methods that distinguish between the case of broken data pipelines or changes to feature semantics on the one hand and a change of user behavior on the other hand. This would enable alerting only in the former scenario.}
    %\item[Model effect on fraud behavior]{How can we measure the causal effect that deploying a new model has on the behavior of fraudsters? Will the performance of the model rapidly degrade when used to action?}
    % How can we measure and exploit the causal effect of deploying a new model on the behavior of the fraudsters.
    % How do we deal 
    %\item[Model performance monitoring]{How do we measure model performance over time under concept drift? Can we distinguish whether the production model is degrading or whether the detection problem is getting harder?}
\end{description}

\section{Conclusion}
\label{sec:conclusion}
We presented an operational model of how an anti-fraud department in an e-commerce organization operates. We formulated a list of practical challenges related to fraud detection, and we derived a list of machine learning research topics that are practically relevant and applicable in anti-fraud departments by addressing some of these practical challenges. We summarized the state of the scientific literature in these research topics and formulated open research challenges that we believe to be relevant to the industry for anti-fraud operations. 

By formulating these open challenges, this paper functions as a research agenda with industry practicality in mind. At the same time, this paper aims to enable future  work in fraud detection to embed their methods in the \emph{organizational context} using the operational model presented in this paper.

\bibliographystyle{splncs04.bst}
\bibliography{main}
\end{document}